\documentclass[journal]{IEEEtran}

\usepackage[colorlinks=true, linkcolor=blue,citecolor=blue, urlcolor=blue]{hyperref}
\usepackage{cite}
\usepackage{array}
\usepackage{graphicx}
\usepackage{amsmath}
\usepackage{amsthm}
\usepackage{amssymb}
\usepackage{cleveref}
\usepackage[utf8]{inputenc}
\usepackage{color}
\usepackage{enumitem}
\usepackage[normalem]{ulem}

\newcommand{\R}{{\mathbb R}}

\renewcommand{\d}{{\mathrm d}}
\renewcommand{\div}{{\mathrm{div}}}
\newcommand{\proj}{{\mathrm{proj}}}
\newcommand{\id}{{\mathrm{id}}}

\newcommand{\WFR}{{\mathrm{WFR}}}
\newcommand{\halflife}{{T_{\frac12}}}

\newcommand{\sct}{{\mathrm s}}
\renewcommand{\det}{{\mathrm d}}

\newcommand{\dist}{d}
\newcommand{\const}{{\mathrm{const}}}
\newcommand{\error}{{\mathrm{err}}}
\newcommand{\groundtruth}{{\mathrm{gt}}}

\newcommand{\cps}{\textnormal{cps}}
\newcommand{\Bq}{\textnormal{Bq}}
\newcommand{\sens}{s}
\newcommand{\mass}{m}
\newcommand{\tn}[1]{\textnormal{#1}}
\newcommand{\prox}{\mathrm{Prox}}
\DeclareMathOperator*{\argmin}{arg min}

\newcommand{\iter}[2]{#1^{(#2)}}
\newcommand{\iterl}[1]{#1^{(\ell)}}
\newcommand{\iterll}[1]{#1^{(\ell+1)}}

\newcommand{\iterz}[1]{#1^{(0)}}

\allowdisplaybreaks % allow line breaks in series of equations

\ifCLASSOPTIONcompsoc
  \usepackage[caption=false,font=normalsize,labelfont=sf,textfont=sf]{subfig}
\else
  \usepackage[caption=false,font=footnotesize]{subfig}
\fi

\begin{document}
\title{Dynamic Cell Imaging in PET with Optimal Transport Regularization}

\author{Bernhard Schmitzer, %
		Klaus P. Sch\"afers, %
        and~Benedikt Wirth% <-this % stops a space
    \thanks{This work was supported by the Deutsche Forschungsgemeinschaft (DFG, German Research Foundation) via Germany's Excellence Strategy through the Clusters of Excellence ``Cells-in-Motion'' (EXC 1003) and ``Mathematics M\"unster: Dynamics -- Geometry -- Structure'' (EXC 2044) at the University of M\"unster.
    B.W.\ was supported by the Alfried Krupp Prize for Young University Teachers awarded by the Alfried Krupp von Bohlen und Halbach-Stiftung.}%
    \thanks{Bernhard Schmitzer is with the Department of Mathematics, Technical University of Munich, Boltzmannstra\ss e 3, 85748 Garching, Germany}% 
    \thanks{Klaus Sch\"afers is with the European Institute for Molecular Imaging, University of Münster, Waldeyerstra\ss e 15, 48149 Münster, Germany.}% 
    \thanks{Benedikt Wirth is with the Department of Mathematics and Computer Science, University of Münster, Einsteinstra\ss e 62, 48149 Münster, Germany.}% 
}

\maketitle

\begin{abstract}
    
We propose a novel dynamic image reconstruction method from PET
listmode data that could be particularly suited to tracking single or
small numbers of cells. In contrast to conventional PET reconstruction
our method combines the information from all detected events not only
to reconstruct the dynamic evolution of the radionuclide distribution, but also to
improve the reconstruction at each single time
point by enforcing temporal consistency. This is achieved via
optimal transport regularization where in principle, among all possible temporally evolving radionuclide distributions consistent with the PET measurement, 
the one is chosen with least kinetic motion energy. The reconstruction is found by convex
optimization so that there is no dependence on the initialization of
the method. We study its behaviour on simulated data of a human PET
system and demonstrate its robustness even in settings with very low
radioactivity. In contrast to previously reported cell tracking algorithms, 
our technique is oblivious to the number of tracked cells. 
Without any additional complexity one or multiple cells can be
reconstructed, and the model automatically determines the number of
particles. For instance, four radiolabelled cells moving 
at a velocity of 3.1 mm/s and a PET recorded count rate of 1.1 cps 
(for each cell) could be simultaneously tracked with a tracking accuracy 
of 5.3 mm inside a simulated human body. 

\end{abstract}


\section{Introduction}
\label{sec:introduction}

Immune and cell-based therapies have become of emerging interest for eradicating or controlling intracellular pathogens that are difficult to treat with conventional therapies \cite{Whiteside2016}. These immunotherapies generally utilize the primary function of the immune system to kill bacterial and viral pathogens or fight cancer cells. Molecular imaging has great potential to detect and identify biomarkers allowing to treat only those patients which most likely respond to a dedicated immunotherapy and, most importantly, to monitor the treatment's success \cite{Ponomarev2017}. 

Treatment planning and control need a quantitative measure of the biomarker's distribution within the patient's body. Positron emission tomography (PET) is a promising molecular imaging technique as it provides quantitative whole-body data at highest molecular sensitivity. The advent of new radiopharmaceuticals based on F-18 or long-lived Zr-89 and Cu-64 (e.g. \cite{Kircher2011, Tavare2016, Bansal2015}) led to the development of new immuno-PET imaging approaches visualizing immune cell behaviour \textit{in-vivo}. 

Besides the use of PET for the visualization and quantitation of biomarkers, a new strategy has been developed recently to follow single cells over time using the excellent sensitivity of PET systems in combination with mathematical modelling techniques (see for instance \cite{Lee2015} and references therein). In these approaches, radio-labelled cells, directly injected or indirectly labelled, are followed over time which is different to standard PET techniques where usually a broad spectrum marker like $^{18}$F-FDG distributes slowly within the body and shows sites of its accumulation. Therefore, the PET information by singular labelled cells is directly related to the process of cellular motion rather than just showing the cell distribution after \textit{long-term} tracer accumulation.
This application is of great interest to understand the underlying biological processes in immunotherapy.  In addition, for the first time, the possibility of tracking the path of individual cells in the body opens up which is of interest to better understand the role and function of circulating tumor cells in the development of cancer metastases.

Note that conventional spatiotemporal PET-reconstruction from listmode data (e.\,g.\ \cite{NiQiAsLe2002,Wa19})
rests upon a (temporally or spatially) smooth parameterization of the radionuclide density (e.\,g.\ via kernel methods or cubic splines)
and typically also a decoupling of space and time parameterization.
Both these features are incompatible with spatiotemporal radionuclide densities induced by individually moving labelled cells
(they could only be utilized if the moving density is spatially smooth and the motion is slow).
Lee et al.~instead proposed a novel algorithm to reconstruct the trajectory of a moving cell directly from PET list-mode data \cite{Lee2015}. The trajectory is modelled as a 3D B-spline function of time, and the mean squared distance between the trajectory and the recorded coincidence events is minimized via non-linear optimization. This approach allows to track single sources in a small animal PET setup with an accuracy below 3\,mm provided that its activity (in Bq) exceeds four times its velocity (in mm/s).
Using the same algorithm, Ouyang et al.\ demonstrated experimentally the feasibility of tracking a moving point source in another small animal PET system based on BGO detectors without intrinsic background radiation \cite{Ouyang2016a}. Due to the reduced noise level, the PET system could recognize moving sources with an activity-to-velocity ratio $> 3.45$\,decay/mm, outperforming the LSO-based PET system.
Langford et al.~were able to track multiple labelled yeast cells by following local maxima in a discretized, smoothed back-projection of the PET signal, albeit at significantly higher activity-to-velocity ratios \cite{Langford-DynPET-2017}.

Our proposed method combines the information from all detected events to reconstruct the dynamic evolution \emph{and simultaneously}
improve the reconstruction quality
by enforcing temporal consistency.
Our approach combines the advantages of \cite{Lee2015,Ouyang2016a} and \cite{Langford-DynPET-2017}. It can track multiple particles, the number of which is automatically determined and does not affect the complexity, at activity-to-velocity-ratios comparable with \cite{Lee2015}. For single particles our functional is closely related to that of \cite{Lee2015}.
We tested the new algorithm on simulated data of a human PET system to evaluate a potential clinical use of PET based cell tracking under realistic conditions.

\section{Mathematical reconstruction model}
\subsection{Measurements and material distribution}
The interior of the PET scanner (the measurement volume) will be denoted by $\Omega \subset \R^3$.
Measurements are performed over a time interval $[0,T)$ which we divide into $M$ subintervals $(\tau_i)_{i=1}^M$ of duration $\Delta T= T/M$ with $\tau_i=[(i-1)\,\Delta T,i\,\Delta T)$.

With $j \in \{1,\ldots,N\}$ we enumerate all scanner detector pairs. The line of response (LOR) $l_j$ of pair $j$ is a straight line through $\Omega$ that connects the centers of the two detectors. In an idealized setting any $\beta$-decay that led to the activation of pair $j$ is assumed to have happened on $l_j$.
More realistically, due to a variety of effects, the $\beta$-decay either happened close to the line or was caused by a random incidence or strong scattering.
The number of detected events in time interval $\tau_i$ and detector pair $j$ is denoted $E_{i,j}$, and we abbreviate
$$E=(E_{i,j})_{i=1,\ldots,M,\,j=1\ldots,N}\,.$$

Our goal will be to reconstruct from the measurement $E$ the corresponding radionuclide density $\rho_t(x)$ at each position $x\in\Omega$ and time $t\in[0,T]$. We write this as
$$\rho=(\rho_t(x))_{t\in[0,T],x\in\Omega}\,.$$
The expected number of decays in time interval $\tau_i$ and a spatial domain $S\subset\Omega$ is given by
\begin{align*}
	\frac{\log 2}{\halflife}\int_{\tau_i} \int_S\rho_t(x)\,\d x\,\d t
\end{align*}
where $\halflife$ denotes the radionuclide halflife.

\subsection{Forward operator}
Each event count $E_{i,j}$ is Poisson distributed with mean $K_{i,j}$ which can be computed from $\rho$ via the forward operator.
In our simplified model we distinguish three different outcomes of a positron decay and subsequent photon emission:
\begin{enumerate}[label=\alph*)]
	\item Detection: The photons undergo at most minor scattering
		before being registered by a pair of detectors.
	\item Scattering: At least one of the photons undergoes substantial scattering which significantly
		alters its direction.
	\item No detection: The emitted photon pair is not detected, e.g.~due to absorption or imperfect scanner sensitivity.
\end{enumerate}
The outcomes of the former two will be produced by linear operators $A^\det_{i,j}$ and $A^\sct_{i,j}$; we set $A_{i,j}=A^\det_{i,j}+A^\sct_{i,j}$ and $K_{i,j} = A_{i,j}\rho$. The latter does not lead to any detected events and thus does not contribute to $K_{i,j}$. Note that for each $(i,j)$, $A_{i,j}$ is a linear function that takes the spatiotemporal material distribution $\rho$ as input and outputs the expected number of detected events in time interval $\tau_i$ and detector pair $j$. We now describe the two operators.

\paragraph{Detection}
We set
\begin{align}
	\label{eq:ADet}
    A^\det_{i,j}  \rho = \frac{\log 2}{\halflife}\int_{\tau_i}\! \int_{\Omega^2}\!\! H_{j}(y)\,G(y,x)\,p^\det\, \rho_t(x)\,\d x\,\d y\,\d t.
\end{align}
$H_j(y)$ encodes the tomography geometry and gives the probability that a collinear unscattered photon pair emitted at $y$ activates the detector pair $j$.
$G(y,x)$ accounts for the positron range and gives the probability that a positron emitted at $x$ annihilates at $y$.
$G$ can also be used to approximately model minor scattering or slight non-collinearity of the photon pair.
We will just consider a spatially homogeneous Gaussian kernel $G(y,x)=\exp(-|y-x|^2/(2\epsilon^2))/(\sqrt{2\pi}\epsilon)^3$ for some fixed $\epsilon>0$.
$p^\det\in[0,1]$ denotes the probability that a photon pair is neither attenuated nor scattered substantially.

Without complicating the approach one could make $p^\det$ space-dependent and use $H_j$ and $G$ to encode spatially varying material properties and scanner sensitivity, however, for simplicity our simulations are performed for constant sensitivity.

\paragraph{Scattering}
For simplicity we assume that scattering changes the photon rays randomly such that the probability of arriving at a detector pair $j$ is homogeneous,
\begin{align*}
    A^\sct_{i,j}  \rho = \frac{1}{N} \frac{\log 2}{\halflife}\int_{\tau_i} \int_\Omega p^\sct\, \rho_t(x)\,\d x\,\d t.
\end{align*}
Similar as above, $p^\sct\in[0,1]$ is the probability for strong scattering.
As with $A^\det_{i,j}$, one can also consider a more elaborate operator that accounts for spatially inhomogeneous scattering. However, we will later see that for our data this approximation is appropriate  (cf.~\cref{fig:simulMixed_offsetDistribution} and corresponding text).
With probability $1-p^\det-p^\sct>0$ photons remain undetected, e.g.~due to attenuation. Random coincidences can be modelled by adding a constant background to $K_{i,j}$.

\subsection{Optimal transport regularization}
Our method aims at a regime of low radiation activity with relatively few detected events.
Reconstructing the particle distribution $\rho$ from the measurements $E$ is therefore a classical underdetermined and ill-posed inverse problem which we must regularize by incorporating additional prior knowledge.
If the radionuclide distribution $\rho$ were temporally almost constant, then by taking long enough time intervals $\tau_i$ one would detect enough decays within $\tau_i$ to reconstruct the sought $\rho$.
However, for quickly changing radionuclide distributions such a framewise reconstruction separately for each time interval $\tau_i$ immediately breaks down.
Instead it becomes necessary to combine information from multiple consecutive time intervals,
of course taking into account that the radionuclide distribution changes between the time intervals.

As $\rho$ describes the physical motion of radioactive particles there will be some temporal consistency, and not all spatiotemporal distributions $\rho$ are equally likely.
In our variational reconstruction approach we model this by a penalty term that prefers physically plausible distributions.
For this we consider the velocity $v_t(x)\in\R^3$ of the radioactive material at each position $x$ and time $t$, and we will penalize too high velocities.
Mathematically, though, it is easier to work with the mass flux $\omega_t(x)=\rho_t(x)v_t(x)$ instead of the velocity, and we abbreviate
$$\omega=(\omega_t(x))_{t\in[0,T],x\in\Omega}\,.$$
The mass flux satisfies the classical mass conservation
\begin{equation}\label{eqn:continuity}
    \partial_t\rho_t(x)+\div\omega_t(x)=0\,.
\end{equation}
Here we assume $T\ll\halflife$ so that the amount of radioactive material does essentially not change over time;
otherwise the constraint could easily be complemented with an additional decay term (leading to so-called unbalanced transport, cf.~\cite{ChPeScVi18}).
In the end we will not only reconstruct $\rho$, but also $\omega$.

To quantify the likelihood of a path $(\rho,\omega)$ we associate with it its physical action
\begin{equation*}
    \mathcal S(\rho,\omega)=\begin{cases}
    \int_0^T\int_\Omega \frac{\|\omega_t\|_2^2}{\rho_t}\,\d x\,\d t&\text{if $\rho\geq0$ and \eqref{eqn:continuity} holds,}\\
    \infty&\text{else.}
    \end{cases}
\end{equation*}
Note that $\omega_t$ can be interpreted as physical momentum of the motion of $\rho_t$,
so $\int_\Omega \tfrac{\|\omega_t\|_2^2}{\rho_t}\,\d x$ is essentially the kinetic energy of all particles at time $t$.
This action is the so-called Benamou--Brenier functional for optimal transport,
and the minimum value that can be achieved for a transport of $\rho_0$ to $\rho_T$ is the squared Wasserstein-2 distance between both measures \cite{BeBr00}.
We will below add $\beta \cdot \mathcal{S}(\rho,\omega)$ as a penalty term to our log-likelihood functional where $\beta \geq 0$ is a weighting parameter.
In essence, we assign a lower penalty to a path $(\rho,\omega)$ the less the mass moves. This acts as a temporal regularization of the particle trajectories (a spatial regularization could be added on top, but since we do not want to impose any further structure on the moving radiotracer distribution this is not done here).
Note that $\mathcal{S}$ is a convex function of $(\rho,\omega)$ and thus amenable to global optimization (see \cite{BeBr00} for details).

\subsection{Penalized log-likelihood functional and unbiasing}
Denoting by $\mathcal P_\lambda(k)=\lambda^k\mathrm e^{-\lambda}/k!$ the Poisson distribution with mean $\lambda$ and writing as before $K_{i,j}=A_{i,j}\rho$, the conditional probability of having the measurement $E$ if the radioactive mass distribution is $\rho$ can be calculated as
\begin{equation*}
    P(E|\rho)
    =\prod_{i=1}^M\prod_{j=1}^N\mathcal P_{K_{i,j}}(E_{i,j})\,.
\end{equation*}
The corresponding negative log-likelihood is
\begin{align*}
	-\log P(E|\rho) & =
    \sum_{i=1}^M\sum_{j=1}^N \left[ K_{i,j}-E_{i,j}\log K_{i,j}\right]
    +\const
\end{align*}
where the terms $\log E_{i,j}!$ are subsumed into a constant.
Inserting $K_{i,j}=A_{i,j}\rho$ and adding the kinetic regularization leads to a penalized maximum likelihood (ML) functional
\begin{align}
	\label{eq:FuncSimple}
    J^E[\rho,\omega]
    =\sum_{i=1}^M \sum_{j=1}^N \left[ A_{i,j}\rho-E_{i,j}\log A_{i,j}\rho \right]
    +\beta\,\mathcal S(\rho,\omega)
\end{align}
to be minimized for $\rho$ and $\omega$.

Unfortunately, the ML estimate in this setting is strongly biased towards declaring every detected photon pair as being unscattered.
Essentially, this is due to $A^\sct_{i,j}\rho$ being very small compared to $A^\det_{i,j} \rho$ in the definition of $A_{i,j}$
(indeed, $A^\sct_{i,j}$ distributes the intensity evenly over all detectors, while $A^\det_{i,j}$ concentrates all the intensity on a few detectors, thus producing much higher intensities there).
This is a well-known deficiency of the ML estimator.
A simple and intuitive way to compensate for this bias is by introducing a tuning parameter $p \geq 0$ that reweighs the contribution of the two forward operators in that term which contains the detections $E_{i,j}$,
\begin{multline}
	\label{eqn:FuncScatter}
    J^{E,p}[\rho,\omega]
    =\sum_{i=1}^M\sum_{j=1}^N A_{i,j}\rho-E_{i,j}\log\big[A_{i,j}^\det\rho+pA_{i,j}^\sct\rho\big]\\
    +\beta\,\mathcal S(\rho,\omega)\,.
\end{multline}
In the following we will assume that events $E_{i,j}$ were most likely caused by unscattered photons if $A_{i,j}^\det\rho>pA_{i,j}^\sct\rho$ and by scattering otherwise.
This interpretation can be justified by a more detailed derivation of the transition from \eqref{eq:FuncSimple} to \eqref{eqn:FuncScatter}.
For the present article the intuition will suffice that the unscattered contribution $A_{i,j}^\det \rho$ dominates for $p$ close to zero, whereas the scattering contribution $pA_{i,j}^\sct\rho$ dominates for very large $p$.
By choosing an intermediate $p$, just the right amount of events will be interpreted as scattered.

Finally, recall that we will be working in the regime of very low radiation activity which means that $E$ is expected to be sparse, i.e.~only a small fraction of entries $E_{i,j}$ will be non-zero.
$\sum_{i=1}^M\sum_{j=1}^N A_{i,j}\rho$ is the expected total number of detected events (with and without scattering). Consequently, there is a function $r : \Omega \to \R$, related to the material properties and the sensitivity of the PET scanner, such that
\begin{align*}
	\sum_{i=1}^M\sum_{j=1}^N A_{i,j}\rho
	= \int_0^T \int_\Omega r(x)\,\rho_t(x)\,\d x\,\d t\,.
\end{align*}
One can then rewrite \eqref{eqn:FuncScatter} as
\begin{multline}
	\label{eq:FuncFinal}
    J^{E,p}[\rho,\omega]
    =\int_0^T \int_\Omega r(x)\,\rho_t(x)\,\d x\,\d t\\
    -\sum_{i=1}^M\sum_{j=1}^N E_{i,j}\log\big[A_{i,j}^\det\rho+pA_{i,j}^\sct\rho\big]
    +\beta\,\mathcal S(\rho,\omega),
\end{multline}
where the sum now only runs over the sparse subset where $E_{i,j}>0$, and only those elements of the full tomography operator must be computed.
In our experiments the area where the particles are moving is small compared to the dimensions of the PET scanner. Thus we may assume that $r$ is spatially constant.
Moreover,
one can show that if $(\rho,\omega)$ minimize $J^{E,p}$ for some $r$ and $\beta$ then $(\rho,\omega)/q$ will minimize $J^{E,p}$ for $q \cdot r$ and $q \cdot \beta $, for any $q>0$. Hence, only the relative scaling of $r$ and $\beta$ is relevant,
and we may for simplicity set $r(x)=1/T$.

\subsection{Relation to single-cell tracking functional}
At first glance our functional looks very different from the tracking functional for a single cell by Lee et al.~in\cite{Lee2015}.
In this section we illustrate that there is a close relation when our functional is constrained to a single particle;
in fact, the functional by Lee et al.\ is a special case of our restricted functional up to slight implementation and modelling variations.
The advantage of our new formulation is the straightforward and easy applicability to tracking of multiple cells (with potentially different unknown amounts of radioactivity).

Let $t \mapsto X(t)$ describe the trajectory of a single particle in $\Omega$; for simplicity assume that $X$ is differentiable such that $X'(t)$ denotes the particle velocity.
The particle mass is denoted by $\mass \geq 0$.
Then the corresponding time-dependent particle distribution $\rho$ and the material flux $\omega$ are
\begin{align*}
	\rho_t & = \mass \cdot \delta_{X(t)}, &
	\omega_t & = X'(t) \cdot \mass \cdot \delta_{X(t)}
\end{align*}
where $\delta_x$ denotes the Dirac delta function centered at $x$.
Thus
\begin{align*}
	\mathcal{S}(\rho,\omega) & = \mass \cdot \int_0^T \|X'(t)\|_2^2\,\d t\,,
\end{align*}
which corresponds to the kinetic energy of the particle integrated over time.
Likewise one computes
\begin{align*}
	A^\det_{i,j} \rho & = \const \cdot p^\det \cdot \mass \cdot \int_{\tau_i} \int_{\Omega} H_{j}(y)\,G(y,X(t))\,\d y\,\d t \\
	\intertext{where we have subsumed some prefactors into a constant. If we further assume that the time intervals $\tau_i$ are very small (such that $X$ is almost constant on $\tau_i$) and that the spatial detector resolution is very high (such that $H_j$ is only non-zero in a very small environment of the LOR $l_j$, which is much smaller than the width of the kernel $G$), then this can be approximated by}
	A^\det_{i,j} \rho & \approx \const \cdot \mass \cdot \int_{l_j} G(y,X(t_i))\,\d y \\
	\intertext{where the constant contains additional new factors and $t_i$ is the midpoint of the time interval $\tau_i$. If we now choose $G(y,x)$ as a Gaussian kernel of width $\epsilon$ one obtains}
	A^\det_{i,j} \rho & \approx \const \cdot \mass \cdot \exp(-\tfrac{\dist_{i,j}^2}{2\epsilon^2})
\end{align*}
where $\dist_{i,j}$ denotes the minimal distance between the point $X(t_i)$ and the LOR $l_j$.

Let us for a moment ignore the contribution of scattering, i.e.~we set $p=0$. If the pair $(\rho,\omega)$ corresponding to a single particle is inserted into \eqref{eq:FuncFinal} one obtains
\begin{multline*}
	J^{E,p}[\rho,\omega] \approx \mass +
	\sum_{i=1}^M\sum_{j=1}^N E_{i,j}\big[\tfrac{\dist_{i,j}^2}{2\epsilon^2}
	-\log \mass \big] \\
    +\beta\,\mass\,\int_0^T \|X'(t)\|_2^2\,\d t
    + \const\,.
\end{multline*}
For fixed $\mass$ the data term in this functional is (up to a factor) the sum of the squared distances $\dist_{i,j}^2$ just as in \cite{Lee2015}. The regularization term $\int_0^T \|X'(t)\|_2^2\,\d t$ leads to the preference of short trajectories with low derivative. While the precise term used in \cite{Lee2015} is different (it penalizes quadratically the change in the cubic B-spline coefficients in a parametrization of $X$) it is very similar in spirit and can in fact be interpreted as a time discretization of $\int_0^T \|X'(t)\|_2^2\,\d t$.

Now consider scattering. One finds $A^\sct_{i,j} \rho = \const \cdot \mass$, thus
\begin{align*}
	-\log(A^\det_{i,j} \rho+pA^\sct_{i,j} \rho) & \approx
	f(d_{i,j}) -\log \mass + \const
\end{align*}
with $f(d) = -\log(\exp(\tfrac{-d^2}{2\epsilon^2})+\tilde{p})$ where $\tilde{p} \geq 0$ is the product of $p$ and some constant.
For $d/\epsilon\ll-\log\tilde p$ one has $f(d) \approx \tfrac{d^2}{2\epsilon^2}$, while for $(d/\epsilon)^2 \gg -\log \tilde{p}$ one finds $f(d) \approx -\log \tilde{p}$. Thus, the function $g(d)=\min\{\tfrac{d^2}{2\epsilon^2},-\log \tilde{p}\}$ can be viewed as a simple approximation of $f$, and indeed the function $g$ is used in \cite{Lee2015} to model scattering and random coincidences.

Summarizing, when restricted to a single particle our functional is indeed very similar to the one proposed in \cite{Lee2015}. Analogously to the above computations one could now derive a generalization of their functional for multiple particles with trajectories given by functions $X_k$ and masses $\mass_k$ for $k=1,\ldots,P$. However, the resulting functional would be severely non-convex in the coordinates $X_k$ and thus difficult to optimize.
The main novelty of our functional is thus not the forward model or the regularization term, which are similar to the ones used in \cite{Lee2015}, but indeed the formulation in terms of time-dependent densities and flux fields, which yields a convex functional that is readily minimized and does not require to manually choose the particle number $P$.

\section{Methods}
\subsection{Discretization and numerical optimization}

The functions $\rho_t$ and $\omega_t$ are discretized by a regular grid of voxels over $\Omega$ for time points $t\in\{0,\Delta T,\ldots,T\}$.
Constraint \eqref{eqn:continuity} and $\mathcal S(\rho,\omega)$ are then discretized using finite differences on a staggered grid as in \cite{PaPeOu14}.
The operators $A_{i,j}^\det$ and $A_{i,j}^\sct$ are discretized by identifying for each pair $j$ of detectors which voxels contribute how much.
In particular the discretization of the reconstruction method is completely independent from the software used to generate the simulated data, thus preventing the fundamental inverse crime \cite{KaSo-InverseCrime-2007}.
Therefore, the obtained results should indeed be meaningful.

Unless stated otherwise, in our experiments we set $T=120\,\textnormal{s}$, $\Omega=[0,L] \times [0,L] \times [0,L/4]$ with $L=160\,\textnormal{mm}$ and discretize $[0,T] \times \Omega$ with a regular Cartesian grid with $65 \times 64 \times 64 \times 16$ points,
resulting in a meshsize of $\Delta L = 2.5\,\textnormal{mm}$.
We set the width $\epsilon$ of the Gaussian kernel $G(y,x)$ from \eqref{eq:ADet} to $\epsilon=5\textnormal{mm}$ which is slightly larger than the value in the simulations (cf.~\cref{fig:simulMixed_offsetDistribution}) to reduce discretization artifacts.

The functional $J^{E,p}$ to be minimized is jointly convex in $(\rho,\omega)$
(indeed, the constraints $\rho\geq0$ and \eqref{eqn:continuity} are linear, and $\mathcal S$ is known to be convex as well \cite{BeBr00}).
Thus, its minimization can be performed via a convex optimization approach.
The optimization problem is then written in the form
\begin{equation*}
    \inf_{(\rho,\omega)} F(K(\rho,\omega))+H(\rho,\omega)
\end{equation*}
where $K$ is a suitable matrix and $F$ and $H$ are convex lower-semicontinuous functionals.
To this optimization problem the primal-dual implicit gradient descent and ascent from \cite{ChPo10} is applied.
The precise form of $F, H$ and $K$ as well as the algorithm are given in the supplementary materials.

\subsection{Framewise reconstruction}
\label{sec:Framewise}
To demonstrate the benefit of temporal regularization we implement framewise reconstruction as a reference,
i.\,e.\ we set $\beta\!=\!0$ and assume $\rho$ to be constant on each interval $\tau_i$. The flow field $\omega$ no longer appears in the functional.
It is trivial to adapt the numerical scheme accordingly.
Apart from our scatter-unbiasing the unregularized functional is then comparable to standard Bayesian reconstruction methods for PET imaging.

We expect that for small $\Delta T$ the framewise method suffers from a lack of signal in each independent time frame. As $\Delta T$ increases, more events become available in each frame, but there will be increasing inaccuracy due to particle motion.

\subsection{Quantification of reconstruction error}
\label{sec:ErrorQuantification}
Using simulated measurements allows us to quantitatively compare our reconstruction results to the ground truth.
We have to simultaneously measure errors in the localization of multiple particles and their respective masses (since PET is quantitative).
Note that both cannot be separated from each other, since incorrectly positioning some mass from $x\in\Omega$ at $y\in\Omega$
can be interpreted both as an error in localizing the mass from $x$ or as an error in reconstructing the correct amount of mass in $x$ and $y$.
In particular we cannot use the squared Euclidean distance from \cite{Lee2015} as it only applies to single particles and cannot take into account masses.
A suitable joint error quantification for particle localization and mass is achieved by the Wasserstein--Fisher--Rao (WFR) metric $d_{\WFR,\alpha}(\rho,\tilde\rho)$ between material distributions $\rho$ and $\tilde\rho$
(see supplementary material or \cite{ChPeScVi18} for a precise definition).
It measures how far on average the material in $\rho$ has to be transported to change $\rho$ into material distribution $\tilde\rho$,
where transport beyond $\text{distance}\approx\alpha$ is replaced by just changing the mass locally (assuming that rather the mass intensity was wrong than its localization).
In detail,
for two Dirac masses $\delta_x$ at $x$ and $\delta_y$ at $y$ one finds
\begin{align*}
	d_{\WFR,\alpha}^2(\delta_x,\delta_y) = 16\,\alpha^2\,\sin^2\big(\min\big\{\tfrac{\|x-y\|}{4\alpha},\tfrac{\pi}{4}\big\}\big)
\end{align*}
where $\alpha$ is a length scale parameter that balances the trade-off between localization and mass errors.
This can be interpreted as a truncated Euclidean distance since $d_{\WFR,\alpha}^2(\delta_x,\delta_y) \approx \|x-y\|^2$ for $\|x-y\|\ll \alpha$.
Consequently, it provides a natural generalization of the Euclidean distance to the setting of multiple particles with different masses.

To quantify our reconstruction errors we pick $\alpha=25\,\textnormal{mm}$ and calculate
\begin{equation*}
    \error^2=\frac{1}{T} \int_0^Td_{\WFR,\alpha}^2(\mathcal{N}(\rho_t),\mathcal{N}(\rho_t^\groundtruth))\,\d t
\end{equation*}
for $\rho^\groundtruth$ the ground truth, $\rho$ our reconstruction, and $\mathcal{N}$ the map that normalizes a non-negative density. We apply normalization to avoid tedious calibration of the absolute scanner sensitivity.
$\error$ can roughly be interpreted as mean localization error for the particles with a truncation for deviations $\gg \alpha$.
It is computed numerically using the code from \cite{Sc16}.

Since $\alpha \gg \Delta L$ (where $\Delta L$ is the distance between two neighbouring grid points) the error inflicted by misplacing all particles by one discrete grid point is approximately
\begin{equation}
	\label{eq:WFRDiscScale}
	\error_{\textnormal{disc}} = \Delta L = 2.5\,\textnormal{mm}.
\end{equation}
This provides a scale for the expected discretization error.

\section{Experimental validation}
\subsection{Monte Carlo Simulation}
\label{sec:simulation}
We work on simulated data.
In this way we can tune the problem difficulty, and reliable ground truth is available, allowing a transparent evaluation of our reconstruction method.

\begin{figure}[bt]
	\centering
	\includegraphics[height=2.6cm]{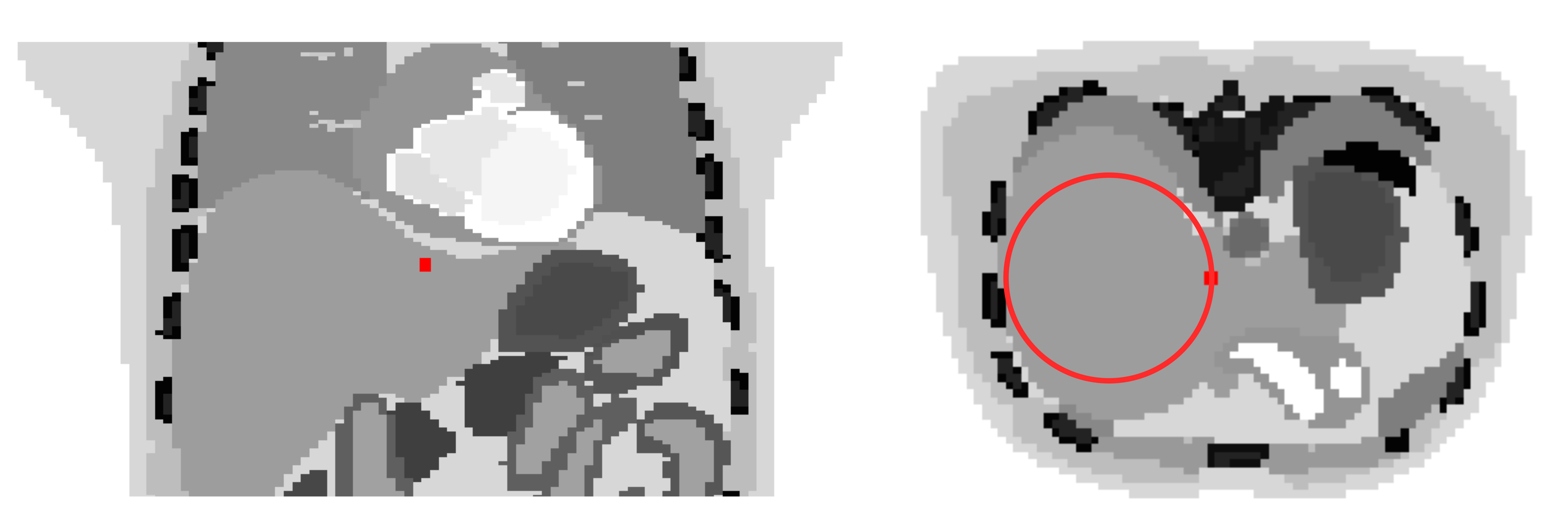}
	\caption{Coronal and transaxial view of the human torso phantom XCAT used for Monte-Carlo simulation of single cell movement. The radio-labelled cell (red dot) is moving along a circular trajectory (red circle) with radius 6\,cm.}
	\label{fig:simulation_coronal_view}
\end{figure}

We simulate list-mode data of a clinical PET scanner with the Geant4 Application for Tomographic Emission (GATE) Monte Carlo tool \cite{Jan2004}. Within this simulation, the geometry and detector concept of the Siemens Biograph mCT PET/CT is implemented to mimic a state-of-the-art human PET system with LSO detector material \cite{Jakoby2011}. To simulate cells under realistic conditions, body anatomy and attenuation is taken from a 4D extended cardiac-torso (XCAT) phantom \cite{Segars2010}. A single cell is simulated as a point source (diameter 1\,mm) with radioactivity of 10\,kBq $^{18}$F ($\halflife = 6586\,\textnormal{s}$).
This cell is moved in 120 steps along a circular trajectory with radius 60\,mm in the x-y plane, see \cref{fig:simulation_coronal_view}. At each step, a Monte-Carlo simulation is performed using 1\,s simulation time, and the information of the detected coincidence events is stored in a list-mode data format.
Note that due to this discretization of a continuous motion the simulated cell moves abruptly in steps of $\pi\,\mathrm{mm}\approx3.14$\,mm which will cause a slight additional error in our reconstructions (so we actually overestimate our reconstruction error in the experiments).

Two different scenarios are simulated. In the first scenario (`without scattering'), $\beta^+$ particles are simulated with typical energies of $^{18}$F followed by positron-electron annihilation inside the human tissue and emission of two 511\,keV gammas. Scattered gamma events are rejected by setting an energy window of 510--650\,keV with no energy blurring.
For a 10\,kBq source activity the rate of counts detected by the scanner was $44\,\cps$ (counts per second), corresponding to a ratio of $\sens=4.4 \cdot 10^{-3}\,\cps/\Bq$.
In the second scenario (`with scattering'), scattered events are partly allowed using an energy window of 435--650\,keV (energy blurring 11.6\,\% at reference energy 511\,keV), a range which is typically used in a clinical setup. The intrinsic natural radiation from Lu-176 within the LSO crystal was not considered in the GATE simulation.
Events were detected with a rate of $104\,\cps$, yielding a ratio of $\sens=1.0 \cdot 10^{-2}\,\cps/\Bq$. The scatter fraction in this scenario was about 18\%.
In both simulations a coincidence window of 4.1\,ns is used.

To obtain more realistic experimentally feasible conditions the activity is then subsampled to rates between 160\,Bq and 800\,Bq in the first simulation scenario (corresponding to 0.7 to 3.6 cps for a single particle) and 500\,Bq in the second scenario (5.2 cps/particle). Multiple particles are simulated by applying a rotation in the x-y plane to the listmode data and then combining the resulting events. This is admissible since the scanner geometry is approximately symmetric around this axis. Different subsamplings of the original data were used so that events from different particles are statistically independent.

To obtain a notion of robustness of our reconstruction method we generated five runs for each simulation scenario. The reconstruction accuracies and scattering ratios reported in the following are the obtained averages, and we visualize the standard deviation among the five runs with errorbars.

\subsection{Basic setup}
Our primary test phantom consists of four cells moving one after another on a circle in the x-y plane with a radius of 60\,mm.
We consider four experimental parameters: the amount $m$ of radionuclides within the cells, the cell velocity $v$, the distance between subsequent cells and whether or not photon scattering is included in the recorded events.
The detection rate (counts per second) is then given by $A=\sens \cdot m \cdot \log 2/\halflife$.
We vary $A$ by varying $m$.
For values of $s$ and $\halflife$ and how $m$ is varied by subsampling the original data, see \cref{sec:simulation}.

Recall that in addition there are two major parameters in the reconstruction model: $\beta$ for kinetic regularization and the scattering tuning parameter $p$.
This setup allows to carefully test the method at different levels of difficulty and to understand how to choose the two reconstruction parameters properly.

\begin{figure}[bt]
	\centering
	\includegraphics[width=8.85cm]{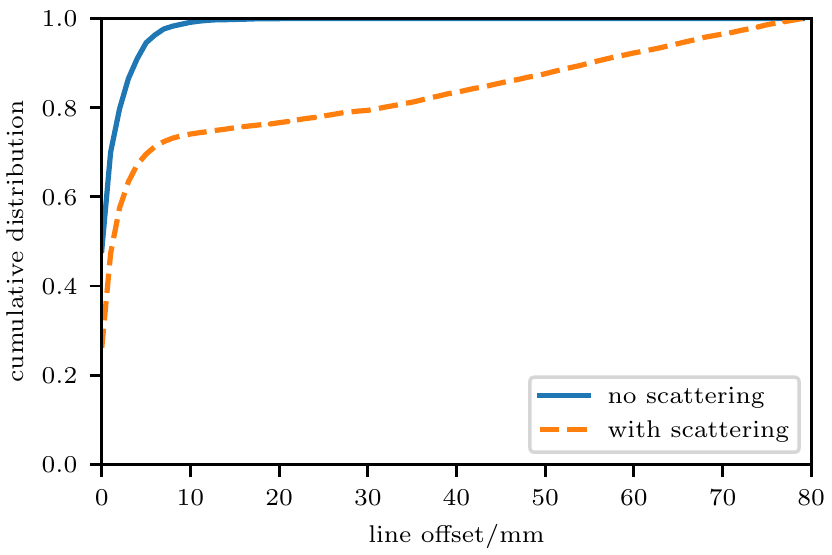}
	\caption{Cumulative distribution of distance between positron emission position and line of response of the activated detector pair for all observed events with and without photon scattering.}
	\label{fig:simulMixed_offsetDistribution}
\end{figure}

\Cref{fig:simulMixed_offsetDistribution} shows the cumulative distributions of the distance from the original positron emission to the line of response of the activated detector pair for all detected events, for simulated data with and without photon scattering.
In absence of scatter this distance is caused by the positron range and photon non-collinearity, as modelled with the kernel $G$ in $A^\det_{i,j}$, see \eqref{eq:ADet}.
With scatter included, minor scattering with small perturbations to the photon orientation is still modelled with $G$.
In events with strong scattering the true photon path will differ substantially from the idealized line of response, and thus the distance between positron emission and line may be much larger.
This is modelled with the operator $A^\sct_{i,j}$, and the long tail of the scattered distribution indicates that the assumption of `uniform' scattering is indeed a reasonable approximation.

\subsection{Temporal regularization}
\begin{figure}[bt]
	\centering
	\includegraphics[]{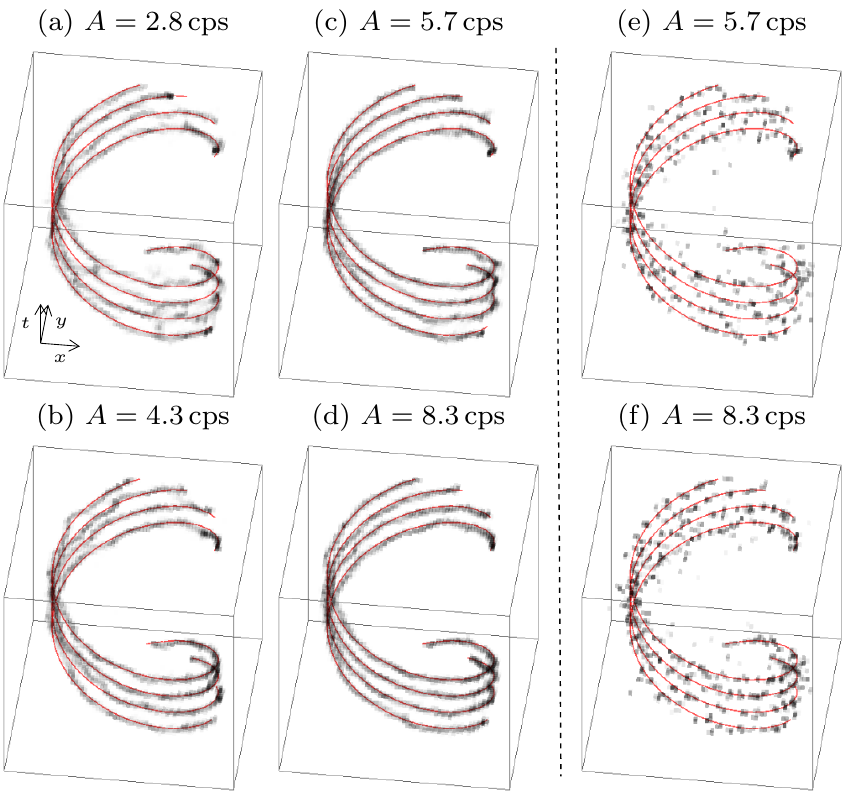}
	\caption{Visualization of $\proj \rho$ (projection of $\rho$ into x-y plane) in $[0,T] \times [0,L]^2$ (axis orientation according to (a)) for different detection rates $A$ (of all particles combined), with temporal regularization (a-d, $\beta=0.9\,\textnormal{s}/\textnormal{mm}^2$) and without (i.e.~framewise, e-f). Red lines indicate ground truth particle positions, dark shading indicates reconstructed particle distributions.
	In the framewise reconstruction there are many spurious artifacts and a lack of temporal smoothness.
	With temporal regularization the trajectories are accurately reconstructed for detection rates as low as 4.3\,cps. Even at 2.8\,cps the particles can be distinguished and their trajectories recognized (see also \cref{fig:vis2DSignal}).}
	\label{fig:vis3DSignal}
\end{figure}

\begin{figure}[bt]
	\centering
	\includegraphics[]{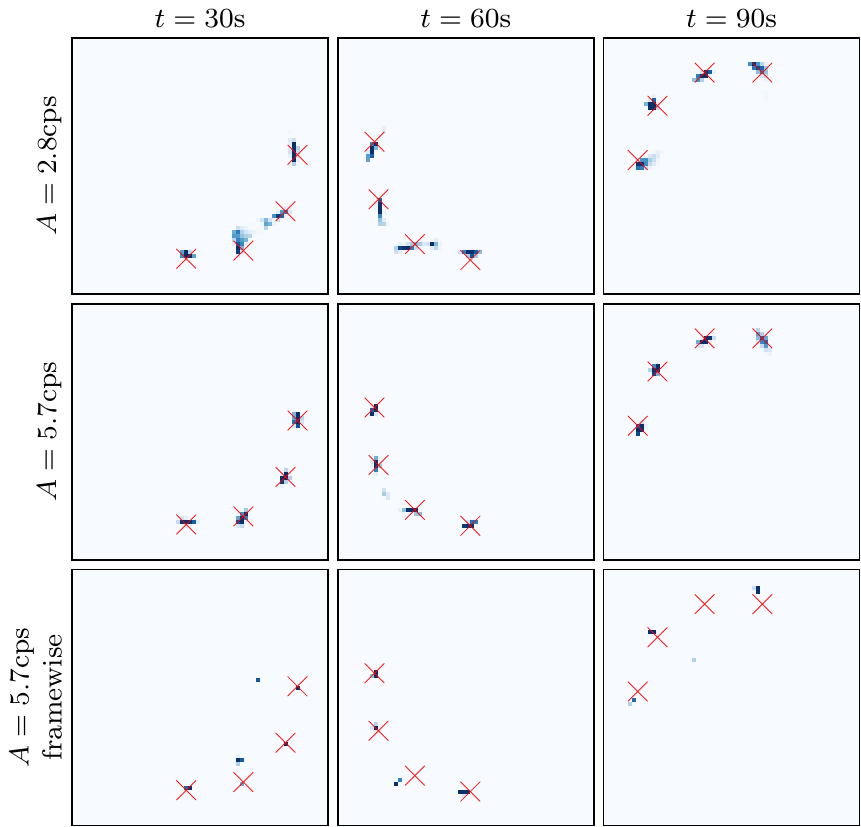}
	\caption{2D slices for some of the visualizations of \cref{fig:vis3DSignal} for various times. Ground truth particle locations are indicated by red crosses. With temporal regularization, at $A=5.6\,\cps$ the particles are reliably located. At $A=2.8\,\cps$ the particles are not as concentrated but can still be separated. Without temporal regularization the reconstruction quality is substantially worse.}
	\label{fig:vis2DSignal}
\end{figure}

We now study the influence of detection rate and temporal regularization.
For simplicity we start without scattering.

Exemplary visualizations of reconstructed trajectories with and without temporal regularization are given in \cref{fig:vis3DSignal,fig:vis2DSignal}. As the particles in the phantom are restricted to the x-y plane, for simplicity we visualize the projection of $\rho$ into this plane (or equivalently, the integration of $\rho$ along the z axis).

\begin{figure}[bt]
	\centering
	\includegraphics[width=8.85cm]{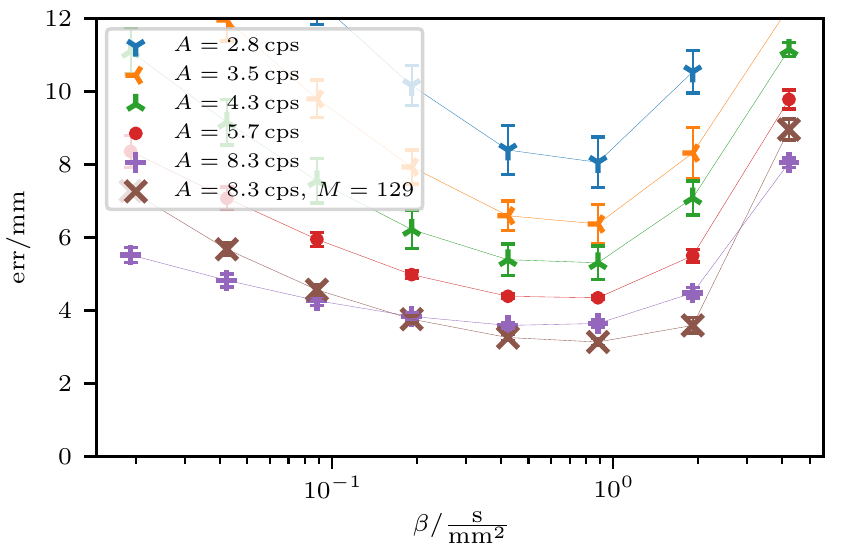}
	\caption{Reconstruction error for different detection rate, temporal resolutions $M$, and $\beta$ for simulations without scattering. %
	The phantom consists of four particles moving one after another on a circle with radius 60\,mm, with velocity 3.14\,mm/s, distance 37\,mm, discretized with $M=65$ time frames (except for the last data series). $A$ is joint detection rate of all four particles. Values are averages over five runs, errors are given by corresponding standard deviation.%
	}
	\label{fig:simul02_WFRvsAKin}
\end{figure}

\Cref{fig:simul02_WFRvsAKin} shows the reconstruction error for different detection rates $A$ and values of $\beta$,
where the detection rate is simply varied by changing the amount of radioactive material in the simulation.
The optimal value of $\beta$ is found to lie between 0.4\,s/mm$^2$ and 1.0\,s/mm$^2$ and is essentially independent of the rate $A$ and number of time frames $M$.
For detection rates of $A=5.7\,\cps$ and above the optimal reconstruction error is almost constant and almost at the level of discretization artifacts (see \eqref{eq:WFRDiscScale}) across about one order of magnitude for $\beta$.
For lower rates the error increases more quickly. The optimal reconstruction error is approximately proportional to the inverse activity (cf.~\cref{fig:simul02_WFRvsAKin_vel} (right)).
For $A=(2.8,4.3,8.3)\,\cps$ the lowest reconstruction errors are approximately $(8.0\pm0.7,5.3\pm0.5,3.6\pm0.1)\,\textnormal{mm}$.
For low regularization $\beta$ the error increases since the coupling between the information contained in different time frames becomes weaker (this effect is more pronounced with more time frames, $M=129$, each of which then contains less information).
As $\beta$ increases the error grows due to overregularization.
Fluctuation between different simulation runs is reasonably small (as the errorbars indicate) showing that our method is robust.
\begin{figure}[bt]
	\centering
	\includegraphics[width=8.85cm]{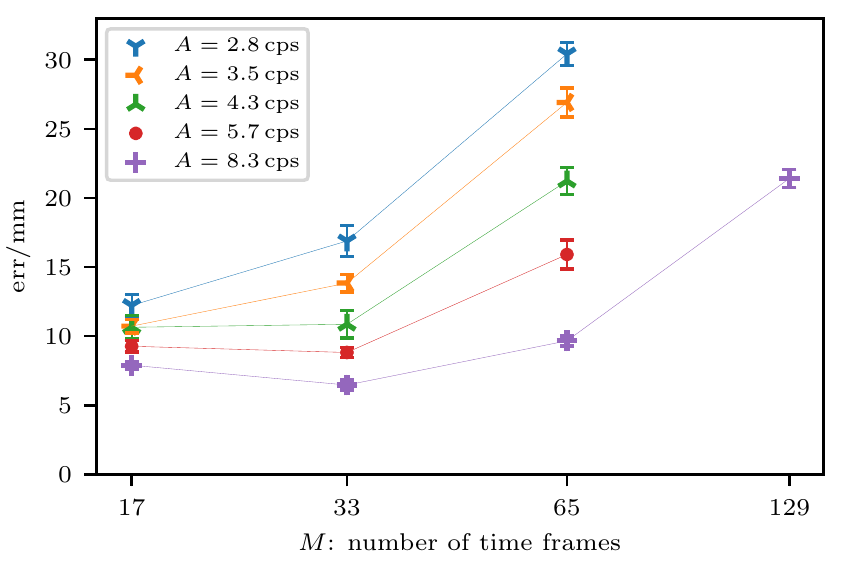}
	\caption{Reconstruction error for different detection rates and temporal resolutions $M$, without temporal regularization. Other phantom parameters are as in \cref{fig:simul02_WFRvsAKin}. Note $y$-axis is in log scale.}
	\label{fig:framewiseErrors}
\end{figure}

For comparison, the reconstruction errors without kinetic regularization (i.e.~framewise reconstruction, \cref{sec:Framewise}) are given in \cref{fig:framewiseErrors}.
The errors are higher than with temporal regularization, and in particular they increase drastically with decreasing detection rate and therefore also with increasing number of time frames (which reduces the available information per frame).
Reducing the number of frames increases the available information per frame but the error eventually increases due to `motion blur'.
The reconstruction with temporal regularization is more robust to little information per frame due to the coupling between the frames.

Parameter $\beta$ balances the strength of temporal regularization versus agreement with the measurements.
Its optimal value depends on the substance amount $m$, particle velocity $v$ and also the halflife $\halflife$ (since our data term uses radioactivity, but the regularization is based on material mass and thus still needs to be scaled by the inverse halflife).
The precise relation can be deduced from the scaling behavior of the functional $J^{E,p}$.

For instance, if one rescales $m \to q \cdot m$, $\halflife \to q \cdot \halflife$ for some $q>0$ (which leaves the rate $A$ and thus the measurement $E$ unchanged) and rescales $\beta \to \beta/q$, then the minimizer of $J^{E,p}$ is rescaled according to $(\rho,\omega) \to (q \cdot \rho, q \cdot \omega)$,
i.\,e.\ the optimal reconstruction is rescaled by the same factor as the true material amount.
As a consequence the reconstruction quality is invariant under this rescaling and thus if $\beta$ is the optimal parameter for $(m,v,\halflife)$ then $\beta/q$ is the optimal parameter for $(q \cdot m, v, q \cdot \halflife)$.
With additional similar arguments one can show that the optimal parameter $\beta$ is given by $C/(\halflife\cdot v^2)$ for a suitable constant $C$.
From \cref{fig:simul02_WFRvsAKin} we deduce that the optimal $\beta$ for $v=3.14\,\textnormal{mm}/\textnormal{s}$ and $\halflife=6586\,\textnormal{s}$ is approximately $0.9\,\textnormal{s}/\textnormal{mm}^2$ and therefore $C\approx 600$.
Unless stated otherwise, in all subsequent experiments we set $\beta$ according to this formula.

In summary, one obtains good reconstructions for a large interval of values of $\beta$, and the dependence of the optimal value on the experimental parameters is well understood.
Also, as compared to the framewise reconstruction, the regularized reconstruction quality is more robust at low detection rates.

\subsection{Particle velocity and tracking efficiency}

\begin{figure}[bt]
	\centering
	\includegraphics[width=8.85cm]{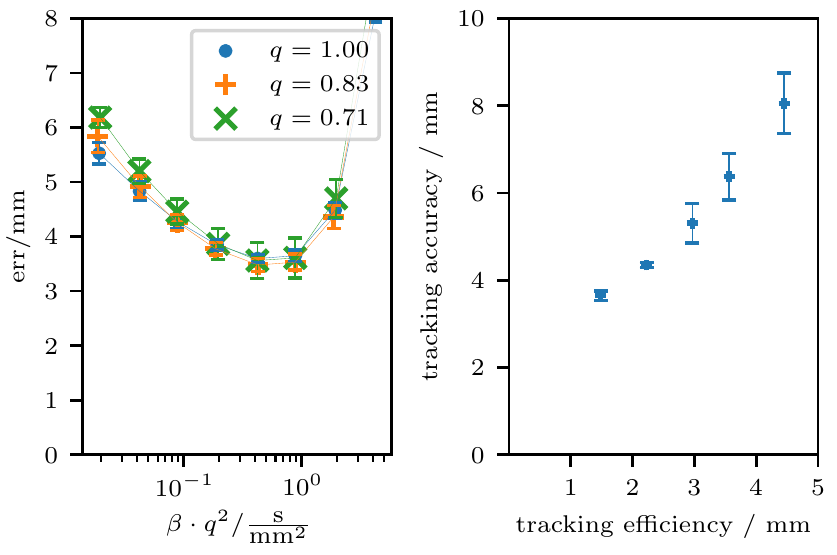}
	\caption{Left: reconstruction error for different particle velocities, detection rates, and $\beta$. Particle velocities $v=q \cdot 3.14\,\textnormal{mm/s}$, detection rate $A=q \cdot 8.3\,\cps$.
	Other parameters as in \cref{fig:simul02_WFRvsAKin}. %
	Right: tracking efficiency vs.~tracking accuracy for the data displayed in \cref{fig:simul02_WFRvsAKin} for $\beta=0.9\,\textnormal{s}/\textnormal{mm}^2$.
	}
	\label{fig:simul02_WFRvsAKin_vel}
\end{figure}

In \cite{Lee2015} it was observed that particle velocity $v$ and detection rate $A$ are not independent parameters with respect to the reconstruction quality.
In essence only their ratio $v/A$ is important which was called \emph{tracking efficiency} in \cite{Lee2015}.
This is intuitive as the tracking efficiency is the average distance between two detected positron decays along the particle trajectory.

This relation can also be observed for our reconstruction functional $J^{E,p}$. Let $q>0$ and rescale $v \to q \cdot v$ and $m \to q \cdot m$ which implies $A \to A \cdot q$ and in particular that $v/A$ remains constant. Then, similar to above, if one rescales $\beta \to \beta/q^2$, one can show that the minimizer of the functional transforms as $\rho_t \to q \cdot \rho_{q\,t}$ and in particular the reconstruction error does not change.
This is confirmed numerically in \cref{fig:simul02_WFRvsAKin_vel} (left). As predicted the curves look very similar.

The relation between tracking efficiency and tracking accuracy for the data of \cref{fig:simul02_WFRvsAKin} is shown in \cref{fig:simul02_WFRvsAKin_vel} (right).
For a detection rate of $A=4.3 \,\cps$ for four particles the detection rate per particle is $A_{\textnormal{single}}=1.1 \,\cps$ which corresponds to a tracking efficiency of $3.0 \,\textnormal{mm}$, at a tracking accuracy of $5.3 \,\textnormal{mm}$.
In \cite{Lee2015} comparable tracking efficiencies are reported at somewhat better accuracies albeit only for a single particle and a smaller scanner geometry providing higher image resolution.

\subsection{Scattering}
\begin{figure}[bt]
	\centering
	\includegraphics[width=8.85cm]{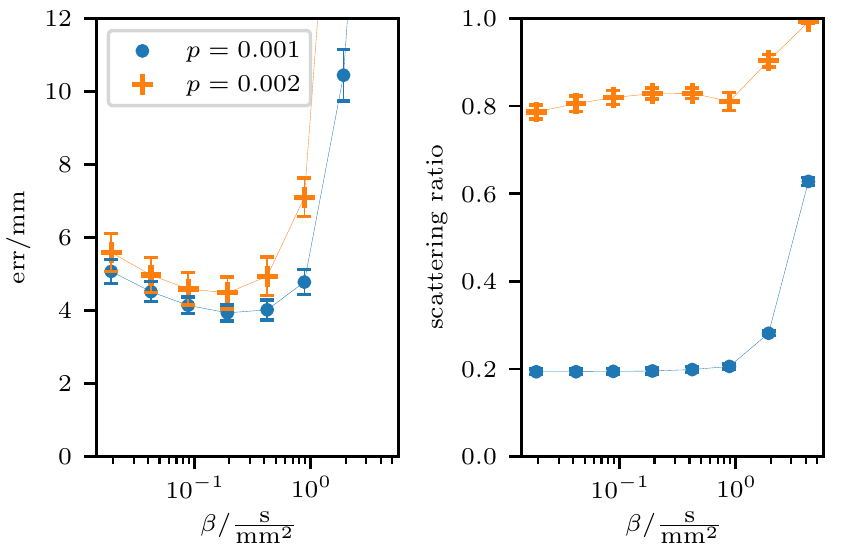}
	\caption{Left: reconstruction error for different $\beta$, $p$ (simulations with scatter). %
	Particle distance 47\,mm, total detection rate (incl.\ scatter) 20.8\,cps, other pa\-rameters as in \cref{fig:simul02_WFRvsAKin}. %
	Right: estimated scattering ratio for same reconstructions.}
	\label{fig:simul03_WFRvsAKin}
\end{figure}

We now turn to data with scattering. We will use the full forward operator involving the parameter $p$ used for unbiasing.

\Cref{fig:simul03_WFRvsAKin} (left) shows the reconstruction error for two values of $p$.
For low and medium values of $\beta$ the behavior is very similar to the no-scattering case (cf.~\cref{fig:simul02_WFRvsAKin}).
However, with scattering the effect of overregularization is more dramatic which is due to a new mechanism that becomes available with scatter modelling: when the kinetic regularization is too strong, the lowest functional values can be obtained by keeping the reconstructed particles almost stationary and declaring almost all detected events as scattering.
\Cref{fig:simul03_WFRvsAKin} (right) shows the corresponding estimated scattering ratios, where we interpret an event $(i,j)$ as `scattered' if $pA_{i,j}^\sct\rho\geq A_{i,j}^\det\rho$.
For small and medium values of $\beta$ the ratio stays almost constant for each $p$, but increases quickly to 1 in the overregularization regime.

For sufficiently low $\beta$ (below overregularization) the scattering ratios for $p=0.001$ and $p=0.002$ are roughly 0.2 and 0.8, respectively. From \cref{fig:simulMixed_offsetDistribution} we see that the former is relatively accurate while the latter is substantially too high and wrongfully discards many unscattered events.
Despite this, the reconstruction error for $p=0.002$ is almost on par with $p=0.001$ since the remaining 20\,\% of undiscarded events provide sufficient information about the particle locations.

Due to the stronger effect of overregularization the optimal value for $\beta$ is slightly reduced. Hence, for scattered data we will set in the following $\beta=C/(\halflife\cdot v^2)$ for $C \approx 130$.

\begin{figure}[bt]
	\centering
	\includegraphics[width=8.85cm]{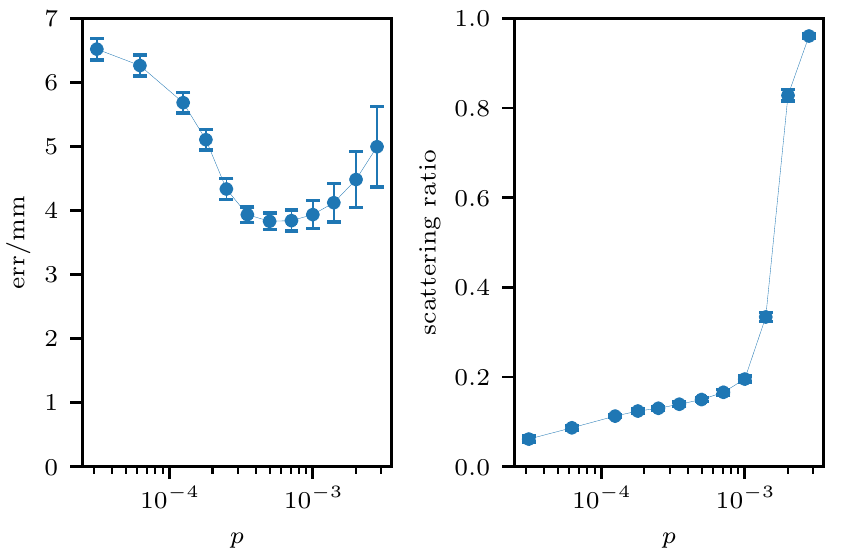}
	\caption{Left: reconstruction error for different $p$ for simulations with scattering. %
	$\beta=0.19$\,s/mm$^2$, other parameters as in \cref{fig:simul03_WFRvsAKin}. %
	Right: estimated scattering ratios for same reconstructions.}
	\label{fig:simul03_WFRvsSigmaB}
\end{figure}

\Cref{fig:simul03_WFRvsSigmaB} illustrates the influence of $p$ on the reconstruction error and estimated scattering ratio in more detail.
The estimated scattering ratio increases monotonically with $p$.
For small $p$ only the `most absurd' events are labeled as scattered. With $p$ the ratio increases, and eventually almost all events are considered scattered.
In accordance, for small $p$ the reconstruction error is higher since some scattered events that are not correctly recognized as such cause artifacts in the reconstruction. For high $p$ the error increases since too many events are discarded as scattered which leaves less reliable information for reconstruction.
The minimum reconstruction error is roughly obtained for that value of $p$ which predicts the correct scatter ratio.
Our experiments show that the optimal value for $p$ does not depend significantly on detection rate or particle velocity, and it is thus relatively easy to calibrate.
If $p^\det$ and $p^\sct$ are known, the nuclide intensity can in principle be reconstructed accurately, but regularization adds a systematic bias towards lower masses. For the chosen optimal values of $\beta$, the reconstructed mass was around 87\% of the true value.

\begin{figure}[bt]
	\centering
	\includegraphics[]{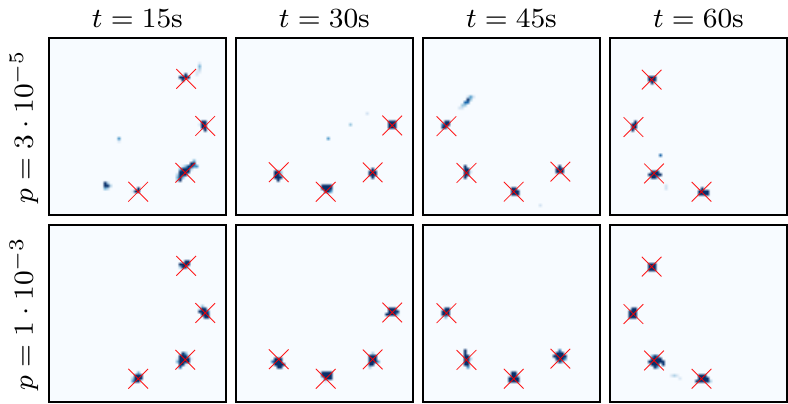}
	\caption{2D slices for two of the reconstructions of \cref{fig:simul03_WFRvsSigmaB} for various times, analogous to \cref{fig:vis2DSignal}. Ground truth particle locations are indicated by red crosses. For $p=3\cdot10^{-5}$ one can see spurious artifacts caused by the failure to identify scattered events. For $p=10^{-3}$ almost all artifacts vanish. Relative to \cref{fig:vis2DSignal} the color scale for the particle densities was magnified as otherwise the scattering artifacts would not be visible.}
	\label{fig:vis2DScattering}
\end{figure}

Reconstructions of data with scattering, with and without proper detection of scattered events are visualized in \cref{fig:vis2DScattering}.

\subsection{Particle distance}
\begin{figure}[bt]
	\centering
	\includegraphics[width=8.85cm]{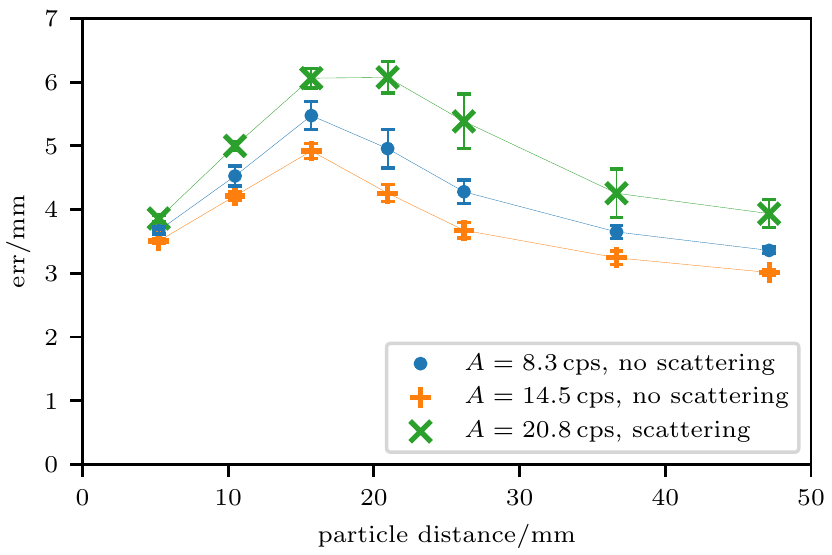}
	\caption{Reconstruction error for varying particle distance for different detection rates, with and without scattering, and optimal $\beta$. All other parameters are as in \cref{fig:simul02_WFRvsAKin} (without scattering) and \cref{fig:simul03_WFRvsAKin} (with scattering).}
	\label{fig:simulMixed_particleDistance}
\end{figure}

In \cref{fig:simulMixed_particleDistance} the reconstruction errors for different detection rates with and without scattering are investigated for varying distances between the particles.
For sufficiently large distances the reconstruction error is low and approximately constant, indicating that particles can be separated cleanly.
As particles come closer the reconstruction error increases. Eventually it decreases again when the particles are essentially merged.
\begin{figure}[bt]
	\centering
	\includegraphics[]{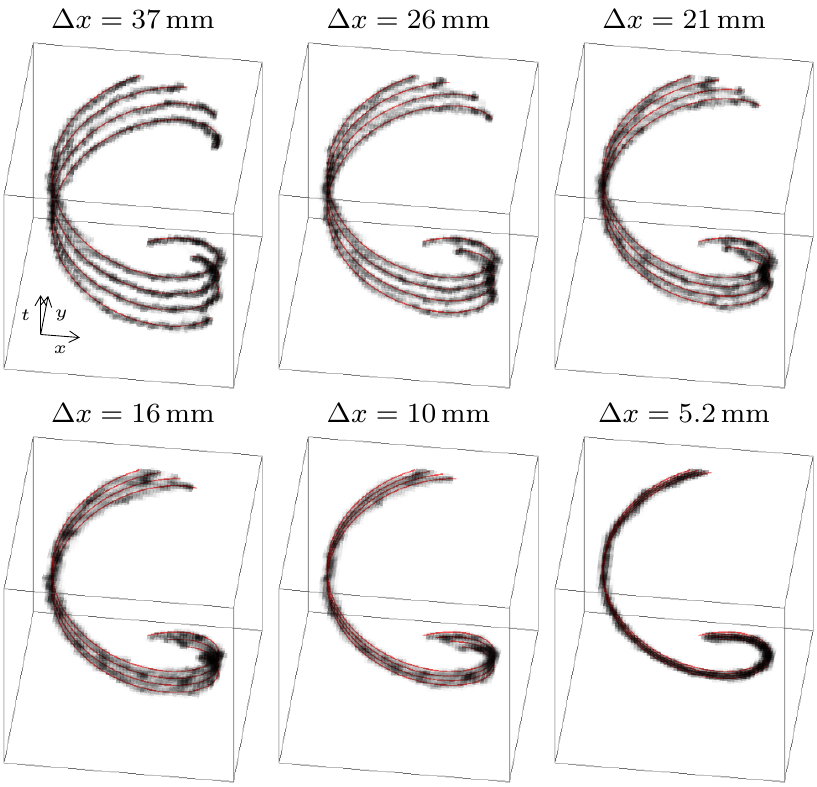}
	\caption{Visualization of $\proj \rho$ analogous to \cref{fig:vis3DSignal} for the data sequence with scattering from \cref{fig:simulMixed_particleDistance} for different particle distances $\Delta x$.}
	\label{fig:vis3DDistance}
\end{figure}
Visualizations of reconstructions for various particle distances for the data of \cref{fig:simulMixed_particleDistance} with scattering are shown in \cref{fig:vis3DDistance}.

\section{Conclusion}
We presented a mathematical model and algorithm for the reconstruction
of a temporally changing radionuclide distribution within a patient
from PET listmode data.
As opposed to classical clinical PET reconstruction,
our method performs a dynamic rather than static imaging, resolving the temporal changes.
It is in particular designed for tracking individual radiolabelled cells,
but could without modification also be applied to the reconstruction of a temporally evolving distribution of injected PET tracers or PET reconstruction under organ (cardiac or respiratory) motion.

In essence, for tracking individual cells via PET there are two existing competing approaches, time binning with independent reconstruction for each bin and path fitting \cite{Lee2015}.
Time binning pretends the radionuclide distribution was stationary over short time intervals
and performs a classical reconstruction over those time intervals.
As a drawback, too large time intervals cause substantial blurring due to the radionuclide motion,
while too small time intervals contain too few detected decays to allow a reliable image reconstruction separately for each time interval.
In contrast, our novel method is essentially continuous in time so that no blurring occurs,
but at the same time exploits temporal consistency
so that the reconstruction almost behaves as if the radionuclide distribution was indeed stationary.
The resulting strong superiority over time binning was demonstrated quantitatively on simulated data.

Path fitting on the other hand tries to fit a particle trajectory to the detected decay events.
While for a single particle this is readily doable,
for multiple particles it introduces the combinatorial problem of deciding which event belongs to which particle.
Our approach can be seen as a relaxation of this method in the sense that
in principle our method also allows particles to split or to merge.
As a consequence our reconstruction boils down to a simple convex optimization.

The above advantages come at the price of two model parameters that have to be adequately picked, however, the influence of either parameter is well-understood.
One parameter determines the amount of detected events
which are interpreted as having undergone scattering (or coming from background noise) and are thus discarded.
Since the amount of scattering and noise is known for each setting
(it depends on the PET scanner, the radionuclide and the imaged object),
this parameter can be easily tuned (once and for all for each setting) by comparing the estimated scatter ratio to the expected one.
The other parameter represents the regularization strength,
and we have shown how to choose it optimally depending on radionuclide halflife and expected particle velocity.

We tested our method on simulated data for imaging a human in a standard clinical PET scanner,
using four cells or particles, each with 160-800\,Bq activity and 3.1\,mm/s velocity
(reconstructions of larger numbers of cells would exhibit the same quantitative behaviour).
For example, at 240\,Bq per cell (corresponding to 1.1\,cps per cell), four cells could be tracked simultaneously with an accuracy of 5.3 mm.
The particle trajectories can clearly be identified and separated up to a distance comparable to the typical positron range (roughly two to five times, depending on the activity).

The chosen cell test setting is highly challenging but not completely unrealistic under the assumption that a single cell can be labelled with an activity of $\sim 200\,\textnormal{Bq}$ per cell. Although the  activity is usually in the range of a few Bq per cell for standard labelling protocols (e.g. \cite{meier2008tracking,faivre201618f}), it was already demonstrated that high activity levels in the range of 50-70 Bq/cell could be achieved using F-18-FDG or Ga-68 labelled cancer cells \cite{Langford-DynPET-2017,Jung2019}.
PET isotopes with short half-life would be preferable for tracking fast moving cells as less molecules need to be attached to the cell providing high activity. Conversely, long living isotopes, such as Zr-89 with a half-life of 78.4\,h, would allow to track cells over several days albeit with limited activity per cell. Thus, the choice of isotope and labelling efficiency will be crucial for tracking accuracy.
Furthermore, moving from a simulated clinical PET scanner to a total-body human scanner, of which prototypes are under evaluation, the sensitivity and thus detection rate could be increased substantially (a 40-fold effective sensitivity was predicted in \cite{cherry2018total}). This means that the radioactivity per cell could be reduced accordingly or particles could be tracked at higher speed, e.g.~also inside large vessels such as the aorta.

Finally, reconstruction results could be further improved
by employing more a priori knowledge.
Here we completely ignored any anatomic information
(which could come from PET-MR or PET-CT systems),
and rather than regularizing the kinetic energy
quantities such as the acceleration might be biologically more relevant.
Also, any time-of-flight (TOF) information of the coincidence events is ignored in our model and its incorporation would require weaker regularization as the reconstruction problem becomes less ill-posed and the signal-to-noise characteristics are substantially improved (as is typically observed in modern PET systems). Thus, the bias introduced by our transport regularization will be reduced, enhancing the tracking accuracy of our algorithm further.

\bibliography{99_bibliography}

%\appendix
\section*{Supplementary material}
\subsection*{Numerical optimization}

We will perform numerical optimization of the discretized problem with the primal-dual implicit gradient descent and ascent from \cite{ChPo10}.
The (primal) variables of the problem are $\rho$ and $\omega$, the mass and flows on the time- and space-staggered grids (cf.~\cite{PaPeOu14}) which we summarize as $x=(\rho,\omega)^\top$.
The discrete problem can then be written as
\begin{align*}
	\inf_{x} F(Kx)+H(x)
\end{align*}
where $K$ is a combination of linear operators given by
\begin{align*}
	K & =\begin{pmatrix}
		Q_{\tn{time}} & 0 \\
		0 & Q_{\tn{space}} \\
		\id & 0 \\
		A & 0
	\end{pmatrix}. \\
\intertext{Here $Q_{\tn{time}}$ and $Q_{\tn{space}}$ are the interpolation operators from the staggered grids to the centered grid \cite{PaPeOu14} and $A=(A_{i,j}^\det+p\,A_{i,j}^\sct)_{i,j}$ is the discretized forward operator (restricted to bins where $E_{i,j}>0$). The functionals $F$ and $H$ are given by}
	F(\rho_{\tn{c}},\omega_{\tn{c}},\tilde{\rho},\mu) & = F_1(\rho_{\tn{c}},\omega_{\tn{c}})+F_2(\tilde{\rho})+F_3(\mu), \\
	F_1(\rho_{\tn{c}},\omega_{\tn{c}}) & = \begin{cases}
			\beta\,\int_0^T\int_\Omega \frac{\|\omega_{c,t}\|_2^2}{\rho_{c,t}}\,\d x\,\d t & \tn{if } \rho_{\tn{c}} \geq 0, \\
			+ \infty & \tn{else,}
			\end{cases} \\
	F_2(\tilde{\rho}) & = \int_0^T \int_\Omega r\,\tilde{\rho}_t\,\d x\,\d t, \\
	F_3(\mu) & = -\sum_{i=1}^M\sum_{j=1}^N E_{i,j}\log \mu_{i,j}, \\
	H(\rho,\omega) & = \begin{cases}
		0 & \tn{if } \partial_t \rho_t + \div\omega_t = 0, \\
		+ \infty & \tn{else.}
		\end{cases}
\end{align*}
Note that $F$ and $H$ are convex and lower-semicontinuous.
The dual variables corresponding to the arguments of $F$ will be denoted by $y=(\varphi_{1,\rho},\varphi_{1,\omega},\varphi_{2},\varphi_{3})$.
Let $\tau, \sigma>0$ with $\tau \cdot \sigma \|K\|^2<1$ where $\|K\|$ denotes the operator norm of $K$.
Then, for initial primal and dual variables $\iterz{x}$ and $\iterz{y}$ the algorithm is given for $\ell=0,1,\ldots$ by
\begin{align*}
	\iterll{y} & = \prox_{\sigma F^\ast}\big(\iterl{y}+ \sigma\,K\,(2\iterl{x}-\iter{x}{\ell-1})\big), \\
	\iterll{x} & = \prox_{\tau H}\big(\iterl{x}-\tau\,K^\top \iterll{y}\big)
\end{align*}
with the convention $\iter{x}{-1}=\iterz{x}$. Here $F^\ast$ denotes the Fenchel--Legendre conjugate of $F$ and $\prox_{\sigma F^\ast}$ denotes the proximal operator of $F^\ast$ with stepsize $\sigma$, defined by
\begin{align*}
	\prox_{\sigma F^\ast}(\tilde{y}) = \argmin_{y} \left\{ \tfrac12 \|y-\tilde{y}\|^2 + \sigma \cdot F^\ast(y) \right\}
\end{align*}
and analogously for $H$. The Fenchel--Legendre conjugate of $F$ decomposes into the conjugates of $F_1$, $F_2$ and $F_3$,
\begin{align*}
	F^\ast(\varphi_{1,\rho},\varphi_{1,\omega},\varphi_{2},\varphi_{3})\!=\!
	F^\ast_1(\varphi_{1,\rho},\varphi_{1,\omega}) + F^\ast_2(\varphi_{2}) + F^\ast_3(\varphi_{3}),
\end{align*}
and so does the proximal operator,
\begin{multline*}
	\prox_{\sigma\,F^\ast}(\varphi_{1,\rho},\varphi_{1,\omega},\varphi_{2},\varphi_{3}) = 
		\big(\prox_{\sigma\,F^\ast_1}(\varphi_{1,\rho},\varphi_{1,\omega}), \\
		\prox_{\sigma\,F^\ast_2}(\varphi_{2}),\prox_{\sigma\,F^\ast_3}(\varphi_{3})\big).
\end{multline*}
Each of these proximal operators is separable into independent operators per grid point or detector bin. The proximal operator $F^\ast_1$ is described in \cite{PaPeOu14}, those for $F^\ast_2$ and $F^\ast_3$ can be determined by elementary calculations. The proximal operator of $H$ amounts to computing the projection onto solutions of the continuity equation which involves the solution of the Poisson equation on a Cartesian grid. This can be performed efficiently with the FFT algorithm \cite{PaPeOu14}.
\subsection*{Wasserstein--Fisher--Rao metric}
The WFR metric between two nonnegative densities $\mu,\nu:\Omega\to[0,\infty)$ is defined via
\begin{multline*}
    d_{\WFR,\alpha}^2(\mu,\nu)
    =\inf\left\{
    	\int_0^1\int_\Omega
    	\left[
    	\frac{\|\omega_t\|_2^2}{\rho_t}
    	+\alpha^2\,\frac{\zeta_t^2}{\rho_t}
    	\right]
    	\d x\,\d t \right| \\
    \rho_t:\Omega\to[0,\infty),
    \omega_t:\Omega\to\R^3,
    \zeta_t:\Omega\to\R
    \text{ for }t\in[0,1],\\
    \left.\vphantom{\int_0^1\left(\frac{\d\omega_t}{\d\rho_t}\right)^2}
    \begin{pmatrix} \rho_0=\mu \\ \rho_1=\nu \end{pmatrix}\!
    \text{and }\!\partial_t\rho_t+\div\omega_t\!=\!\zeta_t
    % \text{ in distributional sense}
    \right\}.
\end{multline*}
Above, the optimization variables are the temporally changing material density $\rho_t$,
the temporally changing vector field $\omega_t$ describing a material transport, and $\zeta_t$ describing a temporal change of material mass.
For simplicity, in our notation we treated $\rho_t$, $\omega_t$, and $\zeta_t$ as functions,
but in the mathematically rigorous definition the minimization is performed in the space of Radon measures,
i.e.\ $\rho$, $\omega$, and $\zeta$ are all measures on $[0,T]\times\Omega$
(for details see \cite{ChPeScVi18}).

By definition $d_{\WFR,\alpha}^2(\mu,\nu)$ is the minimum action to move $\mu$ onto $\nu$, where in contrast to $\mathcal S$ the action now is complemented with a term for mass changes during transport.
The parameter $\alpha>0$ controls how much mass change is penalized compared to transport.
It can be interpreted as the length scale across which a mass misplacement can be interpreted as a localization error
(in fact, it is known that mass transport only happens up to a distance $\pi \cdot \alpha$ in $d_{\WFR,\alpha}$).

For two Dirac measures at $x$ and $y$ with masses $m_x$ and $m_y$ one obtains
\begin{multline*}
	\d_{\WFR,\alpha}^2(m_x \cdot \delta_x,m_y \cdot \delta_y) = \\
	4\alpha^2\,\big(m_x+m_y-2\sqrt{m_x\,m_y} \cos\big(\min\{\tfrac{\|x-y\|}{2\alpha},\tfrac{\pi}{2}\}\big)\big)\,.
\end{multline*}

\end{document}